\documentclass[letterpaper]{article}
\usepackage{aaai}
\usepackage{times}
\usepackage{graphicx}
\usepackage{latexsym}

\newcommand{\nina}[1]{{#1}}

\begin{document}

\newtheorem{theorem}{Theorem}
\newtheorem{lemma}{Lemma}
\newtheorem{definition}{Definition}
\newtheorem{myexample}{Example}
\newtheorem{mytheorem}{Theorem}
\newcommand{\myproof}{\noindent {\bf Proof:\ \ }}
\newcommand{\myqed}{\mbox{$\Box$}}
\newcommand{\nmax}{N}
\newcommand{\zuck}{\mbox{\sc Reverse}}
\newcommand{\lslg}{\mbox{\sc Largest Fit}}
\newcommand{\lsla}{\mbox{\sc Average Fit}}
\newcommand{\lslgbf}{\mbox{\bf Largest Fit}}
\newcommand{\lslabf}{\mbox{\bf Average Fit}}
\newcommand{\myOmit}[1]{}

\title{Complexity of and Algorithms for Borda Manipulation}


\author{
Jessica Davies\\ University of Toronto\\ Toronto, Canada\\ jdavies@cs.toronto.edu  \And
George Katsirelos\\ LRI, Universit\'e Paris Sud 11\\ Paris, France\\ gkatsi@gmail.com \And
Nina Narodytska\\NICTA and UNSW\\ Sydney, Australia\\ ninan@cse.unsw.edu.au \And 
Toby Walsh\\ NICTA and UNSW\\ Sydney, Australia\\ toby.walsh@nicta.com.au}

\maketitle

\begin{abstract}
We prove that it is NP-hard for a coalition of
two manipulators to compute how to manipulate
the Borda voting rule. This resolves one of the last open problems in
the computational complexity of manipulating
common voting rules. Because of this NP-hardness,
we treat computing a manipulation as an approximation
problem where we try to minimize the number
of manipulators. Based on ideas from
bin packing and multiprocessor scheduling,
we propose two new approximation methods to compute
manipulations of the Borda rule.
Experiments show that these methods significantly
outperform the previous best known 
approximation method.
We are able to find optimal manipulations in almost all
the randomly generated elections tested. Our results suggest
that, whilst computing a manipulation of the Borda rule by a coalition
is NP-hard, computational complexity may provide only
a weak barrier against manipulation in practice.
\end{abstract}

\section{Introduction}

Voting is a simple mechanism to combine
preferences in multi-agent systems.
Unfortunately, results like those of Gibbrard-Sattertwhaite
prove that most voting rules are manipulable. That is, it may pay
for agents to mis-report their preferences.
One appealing escape from manipulation
is computational complexity
\cite{bartholditoveytrick}.
Whilst a manipulation may exist, perhaps it is computationally
too difficult to find?
Unfortunately,
few voting rules in common use are NP-hard to manipulate
with the addition of weights to votes. 
The small set of voting rules that are NP-hard
to manipulate with unweighted votes includes
single transferable voting, 2nd order Copeland,
ranked pairs (all with a single manipulator), and 
maximin (with two manipulators) 
\cite{stvhard,bartholditoveytrick,xzpcrijcai09}.

Borda is probably the only
commonly used voting rule
where the computational complexity
of unweighted manipulation remains open.
\citeauthor{xiaEC10} \shortcite{xiaEC10} observe that:
\begin{quote}
{\em ``The exact complexity of the problem [manipulation by a
  coalition with unweighted votes]
is now known with respect to almost all of the prominent voting rules, with the glaring
exception of Borda''}
\end{quote}
It is known that computing a manipulation of Borda is NP-hard
when votes are weighted \cite{csljac20m07},
and polynomial when votes are unweighted
and there is just a single manipulator \cite{bartholditoveytrick}.
With a coalition of manipulators and unweighted
votes, it has been
conjectured that the problem is NP-hard \cite{zuckermanSODA08}.

One of our most important contributions is to close
this question. We prove that computing a manipulation
of Borda with just two manipulators is NP-hard. As a consequence,
we treat computing a manipulation as an approximation
problem in which we try to minimize the number
of manipulators required.
We propose two new approximation
methods. These methods are based on intuitions from bin packing and multiprocessor
scheduling.
Experiments show that these methods significantly
outperform the previous best known approximation method.
They find optimal manipulations in the vast majority
of the randomly generated elections tested.

\section{Background}

The Borda rule is a scoring rule proposed by
Jean-Charles de Borda in 1770.  Each voter
ranks the $m$ candidates. A candidate
receives a score of $m-k$ for appearing
in $k^{th}$ place.  The candidate with the highest aggregated score
wins the election. As is common in the literature,
we will break ties in favour of the coalition of the manipulators.
The Borda rule is used in parliamentary elections in Slovenia and,
in modified form, in elections within the Pacific Island states of
Kiribati and Nauru.
The Borda rule or \nina{modifications of it
are also used by many organizations and competitions including
the Robocup autonomous robot soccer competition, the X.Org Foundation,
the Eurovision song contest, 
anf in the election of the Most Valuable 
Player in major league baseball. 
}
The Borda rule has many
good features. For instance, it 
never elects the Condorcet loser (a candidate
that loses to all others in a majority of head to head elections).
However, it may not elect the Condorcet winner (a candidate that
beats all others in a majority of head to head elections).

We will number candidates from 1 to $m$.
We suppose a coalition of $n$ agents
are collectively trying to manipulate a Borda election
to ensure a preferred candidate $d$ wins.
We let $s(i)$ be the score candidate $i$ receives from
the votes cast so far. A \emph{score vector} $\langle s(1), \ldots,
s(m)\rangle$ gives the scores of the candidates from a
set of votes.
Given a set of votes, we define the \emph{gap} of candidate $i$
as $g(i) = s(d) + n(m-1) - s(i)$.
For $d$
to win, we need additional votes that
add a score to candidate $i$ which is less than
or equal to $g(i)$. Note that if $g(i)$ is negative for any $i$,
then $d$ cannot win and manipulation is impossible.

Our NP-hardness proof uses a reduction from
a specialized permutation problem
that is strongly NP-complete \cite{yujour}.

\begin{definition}[Permutation Sum]
Given $n$ integers $X_1 \leq \ldots \leq X_n$
where $\sum_{i=1}^n X_i = n(n+1)$,
do there exist two permutations $\sigma$ and $\pi$ of
1 to $n$ such that ${\sigma(i)} + {\pi(i)} = X_i$?
\end{definition}

One of our main contributions is to
prove that computing a manipulation
of the Borda rule is NP-hard, settling
an open problem in computational social choice.

\section{Complexity of manipulation}
The {\em unweighted coalition
manipulation problem} (UCM) is to
decide if there exist votes for a coalition
of unweighted manipulators so that a
given candidate wins. As in
\cite{bartholditoveytrick},
we suppose that the manipulators
have complete knowledge about the scores
given to the candidates from the votes of the non-manipulators.

\begin{theorem}
Unweighted coalition manipulation for the
Borda rule is NP-complete with two
manipulators.
\end{theorem}
\myproof
Clearly the problem is in NP. A polynomial witness is
simply the votes that the manipulators cast which
make the chosen candidate win.

To show NP-hardness,
we reduce a Permutation Sum problem
over $n$ integers, $X_1$ to $X_n$, to a manipulation problem with $n+3$ candidates.
By Lemma 1, we can construct an election
in which the
non-manipulators cast votes to give the score vector:
$$\langle C, 2(n + 2) -X_1 + C,\ldots ,2(n + 2) - X_n + C,2(n+2)+C,y \rangle$$
where $C$ is a constant and $y\leq C$. We claim that two manipulators can make candidate $1$ win such
an election
iff the Permutation Sum problem has a solution.

($\Rightarrow$)
Suppose we have two permutations $\sigma$ and $\pi$ of
$1$ to $n$ with $\sigma(i)+\pi(i)=X_i$. We construct
two manipulating votes which have the scores:
$$\langle n+2, \sigma(1), \ldots, \sigma(n),0,n+1\rangle$$
$$\langle n+2, \pi(1), \ldots, \pi(n),0,n+1\rangle$$
Since $\sigma(i)+\pi(i)=X_i$, these give a total score vector:
$$\langle 2(n+2)+C, 2(n + 2) + C,\ldots ,2(n + 2) + C,2(n+1) + y \rangle$$
As $y \leq C$ and we tie-break in  favour of the manipulators, candidate $1$ wins.

($\Leftarrow$)
Suppose we have a successful manipulation. To ensure candidate
$1$ beats candidate $n+2$, both manipulators must put candidate $1$ in first
place. Similarly, both manipulators must put candidate $n+2$ in last
place otherwise candidate $n+2$ will 
\nina{will beat our preferred candidate.}
Hence the final score of candidate $1$ is $2(n+2)+C$.
The gap between the
final score of candidate $1$ and the current score of candidate $i+1$ (where
$1 \leq i \leq n$) is $X_i$. The sum of these gaps is $n(n+1)$.
If any candidate $2$ to $n+1$ gets a score of $n+1$
then candidate $1$ will be beaten. Hence, the two scores of $n+1$
have to go to the least dangerous candidate
which is candidate $n+3$.

The votes of the manipulators are thus of the form:
$$\langle n+2, \sigma(1), \ldots, \sigma(n),0,n+1\rangle$$
$$\langle n+2, \pi(1), \ldots, \pi(n),0,n+1\rangle$$
Where $\sigma$ and $\pi$ are two permutations
of $1$ to $n$.
To ensure candidate $1$ beats candidate $j$ for
$j \in [1,n]$, we must have:
$$2(n+2)-X_j + C+ \sigma(j)+\pi(j) \leq 2(n+2)+C$$
Rearranging this gives:
$$\sigma(j)+\pi(j) \leq X_j$$
Since
$\sum_{i=1}^n X_i = n(n+1)$
and $\sum_{i=1}^n \sigma(i) = \sum_{i=1}^n \pi(i) = \frac{n(n+1)}{2}$,
there can be no slack in any of these inequalities.
Hence,
$$\sigma(j)+\pi(j) = X_j$$
That is, we have a solution of
the Permutation Sum problem.

\myqed

Recall that
we have assumed that the manipulators
have complete knowledge about the scores
from the votes of the non-manipulators.
The argument often put forward for such
an assumption is that
partial or probabilistic information about
the votes of the non-manipulators will
add to the computational complexity
of computing a manipulation.

\section{Approximation methods}

NP-hardness only bounds the worst-case complexity
of computing a manipulation.
Given enough manipulators, we can easily make
any candidate win. We consider
next minimizing the number
of manipulators required. 
For example,
\zuck\ is a simple approximation method
proposed to compute Borda manipulations
\cite{zuckermanSODA08}.
The method constructs the vote of each manipulator
in turn: candidate $d$ is put in first
place, and the remaining candidates are put
in reverse order of their current
Borda scores. The method continues
constructing manipulating votes
until $d$ wins.
A long and intricate
argument shows that \zuck\ constructs a manipulation
which uses at most one more manipulator
than is optimal.

\begin{myexample}
Suppose we have 4 candidates, and the
2 non-manipulators have cast votes:
$3>1>2>4$  and $2>3>1>4$.
Then we have the score vector $\langle 3,4,5,0\rangle$.
We use \zuck\ to construct
a manipulation that makes candidate $4$ win.
\zuck\ first constructs the vote: $4>1>2>3$.
The score vector is now $\langle 5,5,5,3\rangle$.
\zuck\ next constructs the vote: $4>1>2>3$.
(It will not matter how ties
between $1$, $2$ and $3$ are broken).
The score vector is now $\langle 7,6,5,6\rangle$.
Finally, \zuck\ constructs the vote: $4>3>2>1$.
The score vector is $\langle 7,7,7,9\rangle$.
Hence, \zuck\ requires 3 manipulating votes to make
candidate $4$ win. As we see later, this is one more
vote than the optimal.
\end{myexample}

\section{Manipulation matrices}

We can view \zuck\ as greedily constructing
a manipulation matrix.
A \emph{manipulation matrix}
is an $n$ by $m$ matrix, $A$
where $A(i,j)=k$ iff the $i$th manipulator
adds a score of $k$ to candidate $j$.
A manipulation matrix has the properties that
each of the $n$ rows is a permutation of 0 to $m-1$,
and the sum
of the $j$th column is less than
or equal to $g(j)$, the maximum
score candidate $j$ can receive without
defeating $d$.
\zuck\ constructs
this matrix row by row.

Our two new approximation
methods break out of the straightjacket of
constructing a manipulation matrix in row wise order.
They take advantage of \nina{an interesting}
result that relaxes the constraint
that each row is a permutation of
0 to $m-1$. This lets us construct
a \emph{relaxed manipulation matrix}.
This is an $n$ by $m$ matrix that
contains $n$ copies of 0 to $m-1$
in which the sum
of the $j$th column is again less than
or equal to $g(j)$. In a relaxed
manipulation matrix, a row can
repeat a number provided other
rows compensate by not having the number
at all.

\begin{theorem}
Suppose there is an $n$ by $m$ relaxed manipulation
matrix $A$. Then there is $n$ by $m$ manipulation
matrix $B$ with the same column sums.
\end{theorem}
\myproof
By  induction on $n$. In the base case, $n=1$ and we just
set $B(1,j)=A(1,j)$ for $j=1\ldots,m$. In the inductive step, we assume the
theorem holds for all relaxed manipulation matrices with
$n-1$ rows. Let $h(i)$ be the sum of the $i$th column
of $A$. We use a perfect matching in a
suitable bipartite graph to construct the
first row of $B$ and then appeal to the induction
hypothesis on an $n-1$ by $m$ relaxed manipulation
matrix constructed by removing the values
in the first row from $A$.

We build a bipartite graph between
the vertices $V_i$ and $W_j$ for $i\in [0,m-1]$
and $j \in [1,m]$. $V_i$ represent
the scores assigned to the first row of $B$,
whilst $W_j$ represent the columns of
$A$ from where these will be taken.
We add the edge $(V_i,W_j)$ to this bipartite graph
for each $i \in [0,m-1]$,$j \in [1,m]$ and $k \in [1,n]$
where $A(k,j)=i$. Note that there can be multiple edges
between any pair of vertices. By construction,  the degree
of each vertex is $n$.

Suppose we take any $U \subseteq \{V_i | i \in [0,m-1]\}$.
By a simple counting argument, the neighborhood of
$U$ must be at least as large as $U$. Hence, the Hall condition
holds and a perfect matching exists \cite{hall}. Consider
an edge $(V_i,W_j)$ in such a perfect matching.
We construct the first row of $B$ by setting
$B(1,j)=i$. As this is a matching, each $i \in [0,m-1]$ occurs
once, and each column is used exactly one time.
We now construct an $n-1$ by $m$
matrix from $A$ by removing one element equal to $B(1,j)$
from each column $j$. By construction, each value $i \in [0,m-1]$
occurs $n-1$ times, and the column sums are now $h(j)-B(1,j)$. Hence
it is a relaxed manipulation matrix. We can therefore appeal
to the induction hypothesis. This gives us an $n$ by $m$
manipulation matrix $B$ with the same column sums as $A$.
\myqed

We can extract from this proof a
polynomial time method to convert a relaxed
manipulation matrix into a manipulation matrix.
Hence, it is enough to propose new approximation
methods that construct \emph{relaxed} manipulation
matrices. This is advantageous for
greedy methods like those proposed here as we
have more flexibility in placing later entries into
good positions in the manipulation matrix.

\section{\lslgbf}

Our first approximation method, \lslg\ is inspired by
bin packing and multiprocessor scheduling.
Constructing an $n$ by $m$ relaxed manipulation
matrix is similar to packing $n$ objects into
$m$ bins with a constraint on the capacity of
the different bins. In fact, the problem is even similar
to scheduling $nm$ unit length jobs on $n$
different processors with a constraint on
the total memory footprint of the $n$ different
jobs running at every clock tick.
Krause et al. have proposed a simple heuristic
for this problem that schedules the unassigned job with
the largest memory requirement to the time step
with the maximum remaining available memory
that has less than $n$ jobs assigned \cite{krause}.
If no time step exists that can accommodate this job,
then the schedule is lengthened by one step.

\lslg\ works in a similar way to construct a relaxed
manipulation matrix. It assigns the largest unallocated
score to the largest gap. More precisely, it first assigns $n$ instances of $m-1$ to
column $d$ of the matrix (since it is best for
the manipulators to put $d$ in first place in their vote).
It then allocates the remaining $(n-1)m$
numbers in reverse order to the columns
corresponding to the candidate
with the current smallest score who has
not yet received $n$ votes from the manipulators.
Unlike \zuck , we do not necessarily
fill the matrix in row wise order.

\begin{myexample}
Consider again the last example.
We start with the score vector $\langle 3,4,5,0\rangle$.
One manipulator alone cannot increase
the score of candidate $4$ enough to
beat $2$ or $3$. Therefore, we need
at least two manipulators. \lslg\
first puts two $3$s in column $4$ of
the relaxed manipulation matrix.  This gives the score
vector $\langle 3,4,5,6\rangle$.
The next largest score is $2$.
\lslg\ puts this into column $1$ as
this has the larger gap. This gives the score
vector $\langle 5,4,5,6\rangle$.
The next largest scores is again $2$.
\lslg\ puts this into column $2$ giving the score
vector $\langle 5,6,5,6\rangle$.
The two
next largest scores are 1. \lslg\ puts
them in columns $1$ and $3$ giving
the score vector $\langle 6,6,6,6\rangle$.
Finally, the two remaining scores of $0$ are put in columns
$2$ and $3$ so all columns contain
two scores. This gives a relaxed manipulation
matrix corresponding to the manipulating
votes: $4>2>1>3$ and $4>1>3>2$.
With these votes, $4$ wins based on the tie-breaking rule.
Unlike \zuck , \lslg\ constructs
the optimal manipulation with just two
manipulators.
\end{myexample}

\section{\lslabf}

Our second approximation method, \lsla\
takes account of both the size of the
gap and the number of scores
still to be added to each column.
If two columns have the same gap, we
want to choose the column that contains
the fewest scores. To achieve this, we look
at the average score required to fill each
gap: that is, the size of the gap divided by the
number of scores still to be added to
the column. \lsla\
puts the largest unassigned score possible
into the column which will accommodate the
largest average score. \lsla\ does not
allocate the largest unassigned score
but the largest such score that will
fit into the gap. This avoids defeating
$d$ where it is not necessary. If two or more
columns can accommodate the
same largest average score, we 
tie-break either arbitrarily or
on the column containing fewest scores.
The latter is more constrained and
worked best experimentally.
However, it is possible to construct
pathological instances on which the former
is better.

\section{Theoretical properties}

We show that \lslg\ is incomparable to \zuck\
\nina{since there exists  
an infinite family of problems on
which \lslg\ finds the optimal manipulation
but \zuck\ does not, and vice versa. Full proofs of Theorems~\ref{t:prop_1}--\ref{t:prop_2}
can be found in~\cite{dknwcomsoc10}.}

\begin{theorem}
\label{t:prop_1}
For any $k$, there exists a problem with $2k+2$ candidates
on which \lslg\ finds the optimal 2 vote manipulation
but \zuck\ finds a 3 vote manipulation.
\end{theorem}
\myproof (Sketch)
Example 1 demonstrates a problem with 4 candidates
on which \lslg\ finds the optimal 2 vote manipulation
but \zuck\ finds a 3 vote manipulation.
We can generalize Example 1 to $2k+2$ candidates.
\myqed

Unlike \zuck , \lslg\ can use
more than one extra manipulator
than is optimal. In fact the number
of extra manipulators used by \lslg\
is not bounded.

\begin{theorem}
\label{t:prop_2}
For any non-zero $k$ divisible by 36,
there exists a problem with 4 candidates
on which \zuck\ finds the optimal $2k$ vote manipulation
but \lslg\ requires at least  $2k + k/9-3$ votes to manipulate the result.
\label{thm:alg2}
\end{theorem}
\myproof (Sketch)
Suppose $2k$ non-manipulators vote
$1>2>3>4$ and we want to find a manipulation
in which candidate $4$ wins. \zuck\ finds
the optimal $2k$ vote manipulation in
which every manipulator votes $4>3>2>1$.
On the other hand,
if we have $2k + k/9-4$ or fewer rows in a relaxed
manipulation matrix then
it is possible to show that \lslg\ will place
scores in one of the first three columns that exceed the
score of candidate $4$. Hence \lslg\ needs
$2k + k/9-3$ or more manipulators. 
\myqed

\lsla\ is also incomparable to \lslg .
Like \zuck , it finds the optimal
manipulations on the elections
in the last proof.
So far we have not found any
instances where \zuck\ performs
better than \lsla .
However, there exist examples on which \lslg\
finds the optimal manipulation
but \lsla\ does not.

\begin{theorem}
There exists instances on which \lslg\ finds
the optimal manipulation but \lsla\ requires
an additional vote.
\end{theorem}
\myproof
(Sketch)
We failed to find a simple example but a computer search
using randomly generated instances gave the following.
Consider an election
in which the non-manipulators wish the last candidate to win, given the score vector:
$$\langle 41,34,30,27,27,26,25,14\rangle$$
On this problem, \lslg\ finds
the optimal manipulation that makes
the final candidate win but \lsla\ requires
an additional vote.
\myqed

\section{Experimental results}

To test the performance of these approximation methods
in practice, we ran some experiments. Our experimental setup
is based on that in~\cite{wecai10}.  We generated
either uniform random votes or votes drawn from a Polya
Eggenberger urn model. 
In the urn model,
votes are drawn from an urn at random, and are placed back into
the urn along with ${b}$ other votes of the same type.  This captures
varying degrees of social homogeneity. We set ${b} = m!$
so that there is a 50\% chance that the second vote is the same as the
first. In both models, we generated between $2^2$ and $2^7$ votes for varying
$m$. We tested 1000 instances at each problem size.
To determine if the returned manipulations are optimal,
we used a simple constraint satisfaction problem.

\begin{figure*}
\centering{
\begin{tabular}{rr|cccc}
 $m$  &  \# Inst.  & $\zuck$ & $\lslg$  &  $\lsla$ & $\lslg$ beat $\lsla$\\
\hline
4 & 2771 & 2611 & 2573 & 2771 & 0 \\
8 & 5893 & 5040 & 5171 & 5852 & 2\\
16 & 5966 & 4579 & 4889 & 5883 & 3\\
32 & 5968 & 4243 & 4817 & 5879 & 1\\
64 &  5962 & 3980 & 4772 & 5864 & 3\\
128 & 5942 & 3897 & 4747 & 5821 & 2\\
\hline
Total & 32502 & 24350 & 26969 & 32070 & 11 \\
\% & & 75 & 83 & 99 & $<$1
\end{tabular}}
\caption{ \label{fig:table1} Number of uniform elections for which each method found an optimal manipulation.}
\end{figure*}

\begin{figure*}
\centering{
\begin{tabular}{rr|cccc}
 $m$  &  \# Inst.  & $\zuck$ & $\lslg$  &  $\lsla$ & $\lslg$ beat $\lsla$\\
\hline
4 & 3929 & 3666 & 2604 & 3929 & 0 \\
8 & 5501 & 4709 & 2755 & 5496 & 0\\
16 & 5502 & 4357 & 2264 & 5477 & 1\\
32 & 5532 & 4004 & 2008 & 5504 & 0\\
64 &  5494 & 3712 & 1815 & 5475 & 0\\
128 & 5571 & 3593 & 1704 & 5565 & 0\\
\hline
Total & 31529 & 24041 & 13150 & 31446 & 1 \\
\% & & 76 & 42 & 99.7 & $<$1
\end{tabular}}
\caption{ \label{fig:table2} Number of urn elections for which each
  method found an optimal manipulation.}
\end{figure*}

\subsection{Uniform Elections}

We were able to find the optimal manipulation in 32502
out of the 32679 distinct
uniform elections within the 1 hour time-out.
Results are shown in Figure~\ref{fig:table1}.
Both $\lslg$ and $\lsla$ provide a significant improvement over
$\zuck$, solving 83\% and 99\% of instances to optimality.
$\zuck$ solves fewer problems to optimality as the number
of candidates increases, while $\lsla$ does not seem to suffer from
this problem as much: $\lsla$ solved all of 4 candidate instances and
98\% of the 128 candidate ones. 
We note that in every one of the 32502 instances, if $\zuck$
found an $n$ vote manipulation either $\lsla$ did too, or $\lsla$ found an
$(n-1)$ vote manipulation.

\subsection{Urn Elections}
We were able to find the optimal manipulation for 31529
out of the 31530 unique urn elections within
the 1 hour time-out.  Figure~\ref{fig:table2} gives
results. $\zuck$ solves about the same proportion of the urn instances
as uniform instances, 76\%.  However, the performance of
$\lslg$ drops significantly. It is much worse than $\zuck$
solving only 42\% of instances to optimality.  We saw similar
pathological behaviour with the correlated votes in the proof of Theorem
\ref{thm:alg2}.  
The good performance of $\lsla$
is maintained.  It found the optimal manipulation on more than
99\% of the instances. 
\nina{It never lost to
$\zuck$ and was only beaten by
$\lslg$
on one instance in our experiments.}

\section{Related problems}
\nina{There exists an interesting connection between the problem of finding  a
coalition of two manipulators for the Borda voting rule and two other problems
in discrete mathematics: the problem of finding a  permutation matrix with
restricted diagonals sums (PMRDS)~\cite{Brunetti08}
and  the problem of finding multi Skolem sequences~\cite{Nordh10}. 
We consider this connection for two reasons.
First, future advances in these adjacent areas
may give insights into new manipulation algorithms 
or into the complexity of manipulation. 
Second, this connection reveals an interesting open case for
Borda manipulation -- Nordh has conjectured that 
it is polynomial when all gaps are distinct.}

A permutation matrix is an $n$ by $n$
Boolean matrix which is obtained from an identity matrix by a
permutation of its columns. Hence, the permutation matrix contains a
single value 1 in each row and each column. Finding a permutation matrix
such that   the sums of its diagonal elements form a given sequence of
numbers $(d_1,\ldots,d_{2n-1})$  is the permutation matrix with restricted
diagonals sums problem. This problem occurs in discrete tomography,
where we need to construct a permutation matrix from its X-rays for each
row, column and diagonal. The X-ray values for each row and column are
one, while the values for the diagonal are represented with the
sequence $(d_1,\ldots,d_{2n-1})$.

\begin{figure*}[htb]  \centering
    \includegraphics[width=0.8\textwidth]{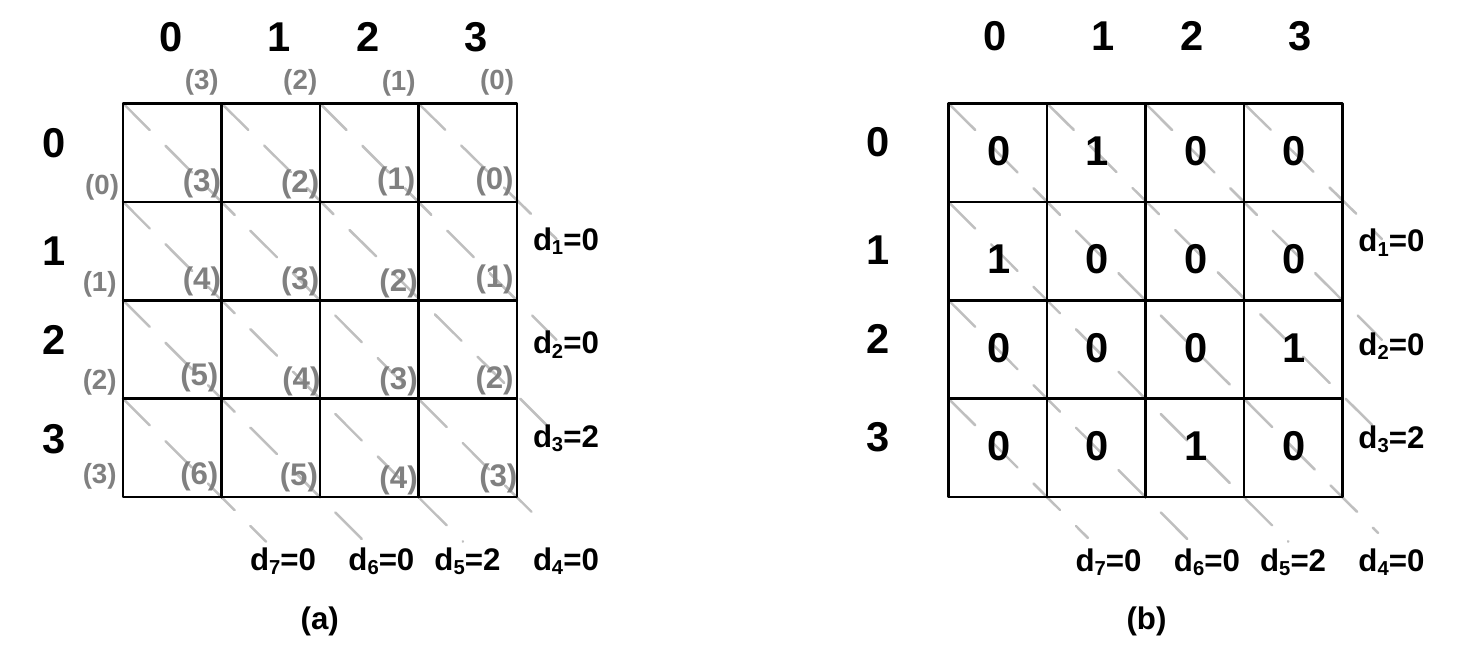}\\
    \caption{(a) A labelling of a permutation matrix; (b) a solution of PMRDS \label{f:xray}}
\end{figure*}

We transform a manipulation problem with  $n$ candidates and $2$
manipulators such that $\sum_{i=1}^n{g_i} = n(n-1)$
to a PMRDS problem. To illustrate the transformation we use the following example
with 5 candidates. Let $\langle 4,4,6,6,0\rangle$ be a score vector, where our favorite candidate
has 0 score,  and $\langle 4,4,2,2\rangle$ be the corresponding gap vector.
We label rows of a permutation matrix with scores of the first manipulator and columns of a
permutation matrix with the reversed scores of the second manipulator.
We label each element of the matrix with the sum of its row  and column labels.
Figure~\ref{f:xray}(a) shows the labelling for our example in gray.

Note that each element on a diagonal is labelled with the same value. Therefore, each
diagonal labelled with value $k$  represents the gap of size $k$ in the
manipulation problem.  Hence, the sum of the diagonal $d_i$ labelled with
$k$ encodes the number of occurrences of gaps  of size $k$.
For example, $d_3=2$ ensures that there are two gaps of size $2$
and $d_5=2$ ensures that there are two gaps of size $4$. The remaining diagonal sums,
$d_i$, $i \in \{1,2,4,6,7\}$,  are fixed to zero.

Consider a solution of PMRDS (Figure~\ref{f:xray}(b)).
Cell $P(0,1)$ contains the value one. Hence, we conclude
that the first manipulator gives the score $0$ and the second
gives the score $2$ to a candidate with the gap $2$.
Similarly, we obtain that the first
manipulator gives the scores $\langle1,3,0,2\rangle$
and the second 
gives the scores $\langle 3,1,2,0\rangle$
to fill gaps $\langle 4,4,2,2\rangle$.
As the number of ones in each diagonal is
equal to the number of occurrences of the corresponding gap, the
constructed two manipulator ballots make our candidate a co-winner.

Finding a coalition of two manipulators for the Borda voting rule is also
connected to the problem of finding multi Skolem sequences used for the construction of Steiner triple system~\cite{Nordh10}.
Given a multiset of positive integers $G$
we need to decide whether there exists a partition of a set $H = \{1,\ldots,2n\}$
into a set of pairs $(h_i,h_i')$, $i=1,\ldots,n$ so that $G \equiv \{h_i-h_i'|h_i,h_i' \in H\}$.
There is a reduction from a manipulation problem with  $n$ candidates and $2$
manipulators such that $\sum_{i=1}^n{g_i} = n(n-1)$
to a special case of multi Skolem sequences with $\sum_{i=1}^nG_i = n^2$
similar to the reduction from
a scheduling problem in~\cite{Nordh10}~\footnote{The reduction implicitly assumes that $\sum_{i=1}^nG_i = n^2$ as the author confirmed in a private communication.}.

\section{Conclusions}

We have proved that it is NP-hard to compute how
to manipulate the Borda rule with just two manipulators.
This resolves one of the last open questions regarding
the computational complexity of unweighted coalition
manipulation for common voting rules. \nina{To 
evaluate whether such computational complexity is important
in practice,} we have proposed two
new approximation methods that try to minimize the
number of manipulators. These methods are based on
ideas from bin packing and multiprocessor scheduling.
We have studied the performance
of these methods both theoretically and empirically.
Our best method finds an optimal manipulation in almost
all of the elections generated.

\section{Acknowledgements}

Jessica Davies is supported 
by the National Research Council of Canada. 
George Katsirelos is supported by 
the ANR UNLOC project ANR 08-BLAN-0289-01.
Nina Narodytska and Toby Walsh 
are supported by the Australian Department 
of Broadband, Communications and the Digital Economy,
the ARC, and the Asian Office of Aerospace 
Research and Development (AOARD-104123).


\vspace{-0.3em}

\section*{Appendix: Constructing votes with target sum}

Our NP-hardness proof requires a technical lemma
that we can construct votes 
with a given target sum.

\begin{lemma}
Given integers $X_1$ to $X_{m}$
there exist votes over $m+1$ candidates
and a constant $C$ such that the final score of
candidate $i$ is $X_i + C$ for $1 \leq i \leq m$
and for candidate $m+1$ is $y$ where $y \leq C$.
\end{lemma}
\myproof
\nina{Our proof is inspired by Theorem 5.1~\cite{xiaEC10}.}
We show how to increase the score
of a candidate by $1$ more than the other
candidates except for the last candidate whose
score increases by $1$ less. 
For instance, suppose we wish to
increase the score of candidate $1$ by $1$ more
than candidates $2$ to $m$ and by $2$ more
than candidate $m+1$.
Consider the following pair of votes:
\begin{eqnarray*}
&1>m+1>2>\ldots>m-1>m \\
&m>m-1>\ldots>2>1>m+1
\end{eqnarray*}
The score of candidate $1$ increases
by $m+1$, of candidates $2$ to $m$
by $m$, and of candidate $m+1$ by
$m-1$. By repeated construction of
such votes, we can achieve the desired result.
\myqed



%

\end{document}